\pdfoutput=1 
\documentclass{article}
\usepackage[utf8]{inputenc}
\usepackage{authblk}
\usepackage{graphicx}
\usepackage{amsmath}
\usepackage{url}
\usepackage[bottom]{footmisc}

\begin{document}

\title{3D U-NetR: Low Dose Computed Tomography Reconstruction via Deep Learning and \\3 Dimensional Convolutions}
\author{Doga Gunduzalp$^{*1}$, Batuhan Cengiz$^{*2}$, Mehmet Ozan Unal$^{2}$, and Isa Yildirim$^{2}$}

\affil{$^{1}$ Technical University of Munich - Department of Electrical and Computer Engineering, Munich, Germany}
\affil{$^{2}$ Istanbul Technical University - Electronics and Communication Engineering Department, Istanbul, Turkey}

\date{}

\maketitle

\textbf{Abstract}: In this paper, we introduced a novel deep learning-based reconstruction technique for low-dose CT imaging using 3 dimensional convolutions to include the sagittal information unlike the existing 2 dimensional networks which exploits correlation only in transverse plane. In the proposed reconstruction technique, sparse and noisy sinograms are back-projected to the image domain with FBP operation, then the denoising process is applied with a U-Net like 3-dimensional network called 3D U-NetR. The proposed network is trained with synthetic and real chest CT images, and 2D U-Net is also trained with the same dataset to show the importance of the third dimension in terms of recovering the fine details. The proposed network shows better quantitative performance on SSIM and PSNR, especially in the real chest CT data. More importantly, 3D U-NetR captures medically critical visual details that cannot be visualized by a 2D network on the reconstruction of real CT images with 1/10 of the normal dose.


\let\thefootnote\relax\footnote{$^{*}$Doga Gunduzalp and Batuhan Cengiz contributed equally to this paper.}

\section{Introduction}

X-ray Computed Tomography (CT) has played a vital role in medicine since its discovery in the 20th century. Unlike X-Ray scans, CT data are volumetric images that are obtained from many 2D projections and allow a view on soft tissue. CT is widely used in the diagnosis of serious illnesses such as cancer, pneumonia, and the epidemic virus Covid-19.

The most traditional technique for reconstruction of a CT image is filtered back projection (FBP) which is based on inverse Radon transform \cite{6499235} and provides sufficient results when full-dose CT is used. However, CT has an inevitable cancer-causing drawback, ionizing radiation. In order to reduce the radiation dose of CT imaging, either the number of projections or the tube current is decreased which results in an ill-posed problem on the reconstruction of an image.

The iterative techniques have been suggested to solve ill-posed problems and reconstruct higher quality images \cite{andersen1984simultaneous,gordon1970algebraic}. Since iterative methods achieve successful results, they are combined with regularization, and regularized iterative methods are proposed. The regularized iterative methods determine a prior knowledge to the image reconstruction problem and one of the traditional prior knowledge for CT image reconstruction is total variation (TV) \cite{sidky2008image}. Besides, there are studies that work on the sinogram domain to improve the quality with regularized iterative models \cite{wang2006penalized}.

In addition, deep learning (DL) models have become a trending solution to inverse imaging along with many optimization problems such as classification \cite{kumar2015lung}, segmentation \cite{skourt2018lung,cciccek20163d} and reconstruction \cite{jin2017deep, zhu2018image,8950464,chen2017low,xie2019deep,adler2018learned,7950488}. Although the DL techniques are based on network training, deep image prior iteratively approaches the inverse image problem by using the randomly-initialized neural network as prior knowledge \cite{lempitsky2018deep,zhu2018image}. As in the regularized iterative models, there are deep learning networks that can operate both in the sinogram and image domain \cite{zhu2018image,8950464,adler2018learned}.

AUTOMAP is a neural network that can achieve mapping in between projection and spatial domains as a data-driven supervised learning task. AUTOMAP is mainly implemented on MRI image reconstruction but it is suggested that it can work on many domain transformations such as CT, PET, and ultrasound \cite{zhu2018image}. Another model that can work from sinogram to image domain is iRadonMAP. iRadonMap achieves improvements on both sinogram and spatial domain alongside the image transformation between domains by implementing the theoretical inverse Radon transform as a deep learning model \cite{8950464}. However, in order to obtain satisfactory results from the networks with fully learned structure, large datasets are needed and in case of insufficient data, they perform worse than FBP and iterative methods \cite{baguer2020computed}. Another network that operates in both projection and spatial domain and gives promising results is the Learned Primal-Dual (PD). Unlike the fully learned networks, Learned PD switches to both sinogram and spatial domains many times during the reconstruction \cite{adler2018learned}.

Recently, the networks operating only in the spatial domain have emerged with the widespread use of autoencoders in medical imaging \cite{7836672}. First, an autoencoder maps the input to the hidden representation. Then, it maps this hidden representation back to the reconstruction vector. Residual connections are widely preferred in denoising and reconstruction networks to model the differences between the input image and ground truth. In addition, overfitting can be prevented and a faster learning process obtained with residual connections. Residual encoder-decoder convolutional neural networks (RED-CNN) model has been proposed by combining autoencoders and residual connections for low-dose CT reconstruction \cite{chen2017low}. Network models designed for one imaging problem can be used for another. Although U-Net is normally created for medical image segmentation \cite{ronneberger2015u}, it is also used for inverse imaging problems \cite{baguer2020computed,jin2017deep}. The image size is reduced by half and the number of feature maps extracted is doubled in each layer at U-Net type networks. The FBP Conv-Net model, which enhances the images obtained with FBP has expanded the coverage of deep learning models in medical imaging. A U-Net like network has been chosen and the modeling of the artifacts created in the sparse view FBP by U-Net is provided with the residual connection which connects the input to the output \cite{jin2017deep}.

Artifacts caused by low projection or low tube current have been greatly denoised with 2D networks mentioned above for the low-dose CT problem. However, in some cases, small details in the sinograms will be lost due to the low dose and it is impossible to reconstruct the missing part from a single sinogram. Since the CT modality has 3D images consisting of multiple 2D image slices, the spatial continuity on the third dimension still exist between slices apart from the continuity in a slice. For this reason, extracting the feature from the adjacent slices is very effective for capturing and enhancing fine details. Liu et. al. mentioned the importance of the third dimension and applied a 1D convolution over 2D convolutions for segmentation of digital breast tomosynthesis and CT \cite{liu20183d}. In addition, Shan et. al. proposed a 3D GAN model via transfer learning from 2D network to indicate the importance of adjacent slice information for low-dose CT \cite{shan20183}.

However, 3D CNNs have also become possible with the increasing availability of computational power and can detect inter-slice features on images. Cicek et al. proposed 3D convolutions for segmentation of CT images \cite{cciccek20163d}. Similar to the RED-CNN network, which works in 2D, Huidong et al. have also taken into account the relationships between 2D CT slices by their network using 3D encoder-decoder structures \cite{xie2019deep}. In this paper, we proposed a U-Net like 3D network called 3D U-NetR (3D U-Net Reconstruction) which is designed to reconstruct low-dose CT images by exploiting the correlation in all three dimensions using 3D convolutions and surface features. The proposed 3D U-NetR architecture has been tested on both synthetic and real chest CT data. In addition, the established experimental setups include both sparse view and low current dose reduction techniques.

\section{General Problems in Low-Dose Computed Tomography}

The CT reconstruction can be expressed as a linear inverse problem as:

\begin{equation}
\label{eq1}
y= Ax + \eta
\end{equation}

where  $A\in{R}^{k\times l}$ represents the forward operator. $x\in{R}^{l}$ is the vector form of the ground truth CT image and $y\in{R}^{k}$ is the vector form of the sinogram. In addition, $\eta\in{R}^{k}$ represents the noise in the system \cite{unal2020unsupervised}.

The number of measurements ($k$) is reduced to obtain CT images from fewer numbers of projections. Therefore the forward operator ($A$) takes the form of a fat matrix and the sparse CT inverse problem occurs with the formation of a non-invertible forward operator. The projections used in sparse CT problems have a high signal-to-noise ratio (SNR) value, but the low number of projections makes inverse operation an ill-posed problem. Another way to reduce the dose is to decrease the signal power by reducing tube current and peak voltage while the number of projections is constant. Any decrease in the signal power, in other words, lower SNR value, is mathematically modeled by increasing the variance of the $\eta$ in \eqref{eq1} and noisy sinograms are obtained. Despite a sufficient number of observations being obtained, each observation has a low SNR value.

The lower frequencies are sampled far more than higher frequencies when a Radon transform is applied to a CT image. Therefore, the traditional methods use inverse Radon transform to solve the low-dose CT problem by first performing a filtering process. Thus, a low-frequency dominant image is reconstructed after filtering. The FBP method applies filters such as Ramp, Hann, Hamming to the sinogram before doing the inverse Radon transform \cite{Pan_2009}.

Iterative and DL-based solutions obtain the measurement results with \eqref{eq1} and calculate the error between measurement and ground truth CT image. The optimization problem of an image to image reconstruction method can be defined as:

\begin{equation}
\label{eq2}
\hat{w} = \operatornamewithlimits{argmin}\limits_{w}\|f_w(X)-Y\|^2
\end{equation}

where $X$ and $Y$ represent the sparse or noisy CT image and the ground truth CT image respectively. In addition, $w$ is the parameters of the model and $f_w$ indicates the nonlinear reconstruction functions such as iterative and DL-based solutions \cite{7950488}. The function whose parameters minimize \eqref{eq2} is considered as the solution.

\section{Proposed Method}

The success of 2D deep learning-based solutions such as FBP-ConvNet \cite{jin2017deep}, RED-CNN \cite{chen2017low}, Learned PD \cite{adler2018learned} and iRadonMAP \cite{8950464} for inverse CT problems has been demonstrated in the literature. However, it is impossible to detect and reconstruct inter-slice detail losses resulting from sparse or noisy views by a 2D network. On the other hand, these details can be recaptured when the correlations between the slices are taken into account. Therefore, it is possible to optimize a reconstruction based on the 3D surface features rather than 2D edge features. Based on this insight, we propose a deep learning-based solution for inverse CT problems called 3D U-NetR which utilizes 3D convolutions and U-Net Architecture. 3D U-NetR operates by mapping initially reconstructed sinograms with FBP to the ground truth volumetric images. The proposed reconstruction process is not limited only to CT images and can be applied to any 3D imaging modality.

\begin{figure*}[!t]
  \begin{center}
  \includegraphics[width=4.8in]{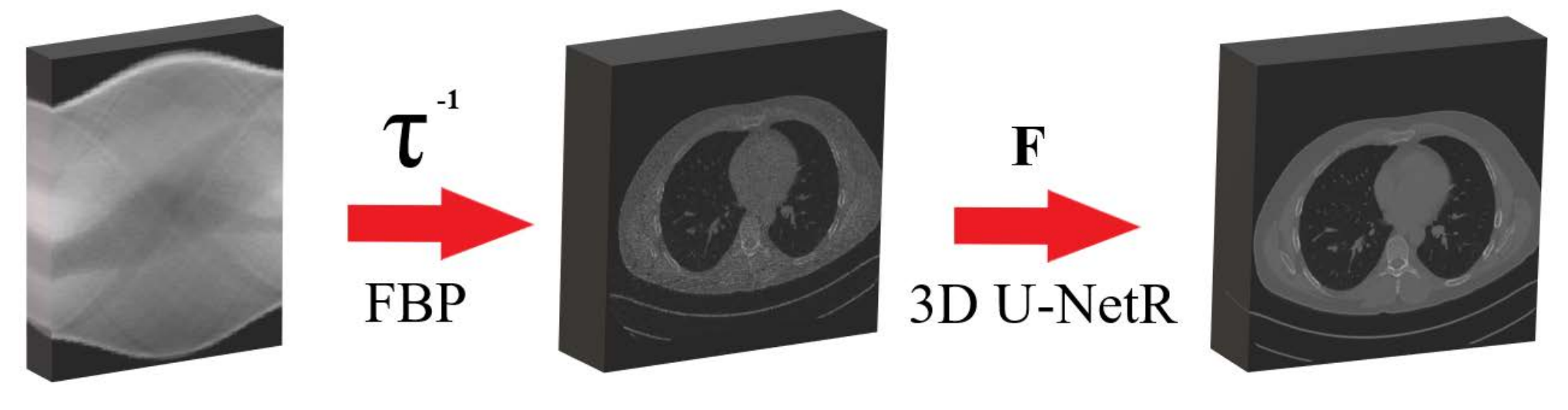}
  \caption{Proposed working schema with 3D U-NetR. $\tau^{-1}$ represents FBP reconstruction of sparse and noisy sinogram and F denotes image to image mapping by the proposed 3D U-NetR.
  }\label{fig:pipline}
  \end{center}
\end{figure*}

Firstly, spatial domain forms of the sparse or noisy sinograms are reconstructed with the inverse operator which can be defined as:

\begin{equation}
\label{eq3}
X = \tau^{-1}(x)
\end{equation}

where $x$ represents the sparse or noisy sinogram of the image and $X$ represents the volumetric low-dose CT image. In addition, $\tau^{-1}$ is the inverse operator and for our case, it is the FBP operator. Low-dose CT images are mapped to ground truth images with minimum error using the 3D U-NetR architecture. This can be expressed as:

\begin{equation}
\label{eq4}
\hat Y = f(X)
\end{equation}

where $\hat Y$ represents the volumetric reconstructed CT image and $f$ is the trained neural network which is 3D U-NetR. The working principle of the proposed reconstruction method is given in Fig.~\ref{fig:pipline}.

\subsection{Network Architecture}

\begin{figure*}
  \begin{center}
  \includegraphics[width=4.8in]{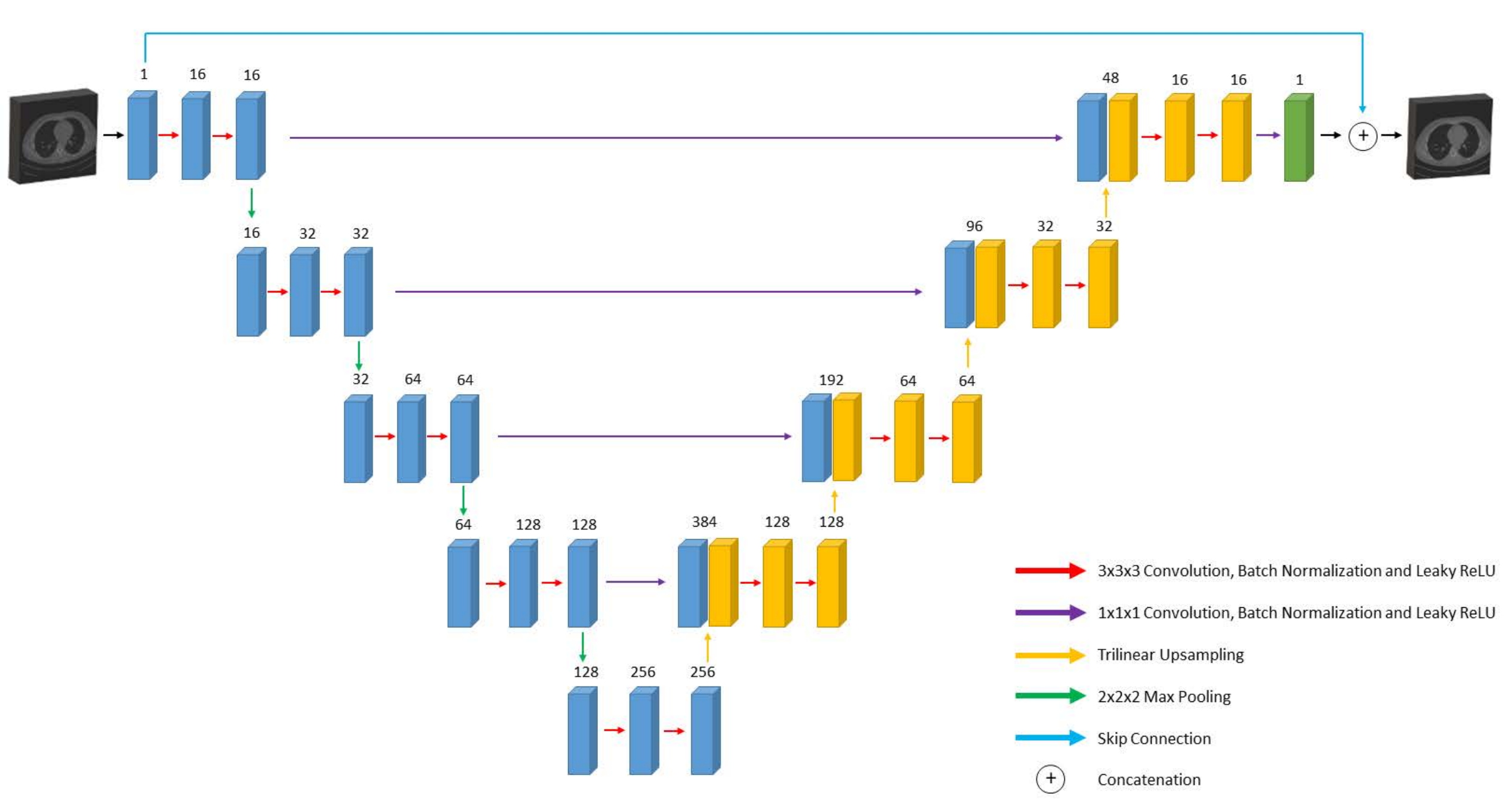}\\
  \caption{Network Architecture of 3D U-NetR.}
  \label{fig:network}
  \end{center}
\end{figure*}

Based on the success of the 2D FBP-ConvNet \cite{jin2017deep} architecture and the 3D U-Net used for segmentation \cite{cciccek20163d}, a U-Net like network is built with 3D CNNs. Fig.~\ref{fig:network} describes the 3D U-NetR architecture. The network is a modified U-Net with 4 depths which can be inspected as analysis and synthesis parts. The analysis part of the network contains 2 blocks of 3$\times$3$\times$3 convolution, batch normalization, and leaky ReLU in each layer. Two layers in consecutive depths are connected with 2$\times$2$\times$2 max-pooling with stride 2.

Starting from the deepest layer, layers are connected with a trilinear interpolation process with scale factor 2 and followed by 2 blocks of 3$\times$3$\times$3 convolution, batch normalization, and leaky ReLU for synthesis. Before the convolution blocks, channels are concatenated with the feature maps from the skip connections of the corresponding analysis layer. Skip connections are used to solve the vanishing gradients problem and carry the high-resolution features. On the other hand, trilinear interpolation is chosen as a simple 3D interpolation method.  Finally, all channel outputs are summed by a 1$\times$1$\times$1 convolution block into 1 channel image and the result is added to the input with shortcut connection.

Overall, the 3D U-NetR architecture contains 5,909,459 parameters which are three times of the 2D structure. The number of layers and filters are kept the same in both networks for a fair comparison which naturally results in an inequality of number of parameters. Due to high number of parameters and memory limitations, the number of filters started from 16 in the first layer and continued to double in each layer up to the deepest one. The number of filters used in the deepest layer thus became 256. In the synthesis part, the number of filters in each layer starting from the deepest layer to the output layer decreases by half in the same manner.

The skip connections contain 1 block of 1$\times$1$\times$1 convolution, batch normalization, and leaky ReLU rather than shortcut connection to be able to tune the number of residually connected channels. In addition, a shortcut connection is connected from the input to the output since the main purpose is to reduce the noise in the FBP images. The random noise modeling of the network is provided with the shortcut connection.

\section{Experimental Setup}

\subsection{Dataset Preparation}

Two datasets are used for the experimentation of the proposed method. Because of the nature of the CT modality and the network architecture, 3D datasets are prepared rather than shuffled 2D CT image slices. Firstly, synthetic data which is a 3D version of the 2D ellipses dataset of Deep Inversion Library (DIV$\alpha l$) is prepared \cite{Dival}. For human CT experiments, a chest dataset acquired from Mayo Clinic for the AAPM Low Dose CT Grand Challenge is used as the real CT dataset \cite{mccollough2020low}.

\subsubsection{Synthetic Data}

The previously used ellipses dataset in the literature \cite{jin2017deep,adler2018learned} which contains randomly generated ellipses is modified to create random ellipsoids in a 3D space. In the ellipses dataset, the number of ellipses in each image slice is selected from a Poisson distribution with an expected value of 40 and limited to 70. For our ellipsoid dataset, the number of ellipsoids in each volume is selected from a Poisson distribution with an expected value of 114 and limited to 200. Later, each volume is normalized by setting all the negative values to zero and dividing them to the maximum value of the volume. Finally, all the volumes are masked with a cylindrical mask along the slice axis in order to be similar to CT images. 

Parallel beams with 60 views and 182 detectors are chosen as projection geometry. A sparse view sinogram of each volumetric image slice is obtained with the forward operator. In addition, additive white Gaussian noise (AWGN) has been applied to the sinograms with 35 dB SNR. Instead of using a signal-dependent noise such as the Poisson distribution, the Gaussian distribution is used to keep the synthetic data simple. The sinograms are reconstructed with a 2D FBP with Hann filter which has 0.8 frequency scaling for each volumetric image slice. 2D FBP is chosen instead of 3D FBP operation to prevent the reconstruction of an extra ellipsoid due to the artifacts in the third dimension.

220 different volumetric images are generated where each image has 128 slices of 128$\times$128 pixel images. Total volumes are separated as 192 training, 8 validation, and 20 test volumes.

\subsubsection{Real Chest CT Data}

The real chest CT dataset from Mayo Clinic consists of full-dose and 1/10 of the normal dose (low-dose) CT image pairs. The 1/10 of the normal dose data are noisy and have full-view images as there is no reduction in the number of projections. The low-dose images are created using a realistic and scanner specific noise insertion model based on Poisson distribution, and inverse square relation between noise and dose is used to calculate the dose level \cite{moen2021low}. Since 2 patients had an unequal number of low-dose and full-dose images, they are excluded from the dataset, which is containing 50 patients in total. 11 patients are excluded to decrease the variance of pixel spacing values since convolution operation is naturally not zoom resistant \cite{Goodfellow-et-al-2016} and different pixel spacing values have the risk of adversely affecting the training. At the point where the artifact is present in the CT images, the pixel values of the image become high, which disrupts the general structure. For this reason, we excluded 9 patients as they contained artifacts. On the other hand, slices are evenly spaced for every patient with a 1.5 mm slice thickness.

Originally, selected patients have 340 ± 18 slices but only the middle 256 slices are used to focus on the middle part of the volumetric image where the main medical information exists. As Luschner et. al. have mentioned, the real CT data contain circular reconstructions, and the data must be cropped at a square inside this circle to prevent value jumps at the circle's boundaries \cite{leuschner2019lodopab}. Accordingly, we focused the middle 384$\times$384 pixels of each slice where there were 512$\times$512 pixels in the original one. Consequently, a dataset of 28 patients with $(0.735\pm0.036)^2$ mm$^2$ pixel spacing, 1.5 mm slice thickness, and 384$\times$384$\times$256 voxels is prepared.

Total 28 patients are separated into 25 training and 3 test volumes. In addition, the volumetric images which are 384$\times$384$\times$256 voxels are divided into 18 volumetric patches with the size of 128$\times$128$\times$128 voxels because of the memory limitations.

\subsection{Training Strategy}

The Tesla T4 graphic processing unit (GPU) with 16 GB memory and GeForce RTX 2080 Ti with 11 GB memory are used during the trainings. Tesla T4 GPU is used in the training with ellipsoids dataset because of the high memory capacity, and GeForce RTX 2080 Ti is preferred in the training with real chest CT dataset due to high processing power. L1 norm is utilized for optimization since it shows better performance in image denoising problems compared to L2 norm \cite{zhao2016loss}. The error minimization between the reconstructed image and ground truth image can be achieved with different algorithms. ADAM optimizer is preferred for this work. The 3D U-NetR architecture is implemented with PyTorch toolbox \cite{pytorch}. A batch consists of 128$\times$128$\times$128 volumetric images in 3D U-NetR training. Therefore, when the batch size is higher than 4, the memory becomes insufficient for both GPUs and the batch size is selected as 4 for the ellipsoids dataset and 3 for the real chest CT dataset for 3D U-NetR training. 0.001 is chosen as the learning rate and the coefficients to be used for finding the mean of the gradients and its square are selected as 0.9 and 0.999 by default. The 3D U-NetR architecture is trained with ellipsoids for 745 epochs, then the real chest CT for 1108 epochs. The training continues until the change in loss values become negligible. Training of ellipsoids and real chest CT images take approximately 62 hours and 166.2 hours, respectively.

The 2D U-Net architecture is also trained with the slices of the same datasets in order to show the contribution of the third dimension. PyTorch, Tesla T4 and GeForce RTX 2080 Ti GPUs, L1 norm loss function, and ADAM optimizer are used as in the 3D U-NetR training. In addition, optimizer parameters such as learning rate, gradient coefficients are kept the same, only batch size is changed. A batch consists of slices that have a 128$\times$128 size and are taken from the volumetric images in 2D U-Net training. The batch size is selected as 384 for the ellipsoids dataset and 256 for the real chest CT dataset. The 2D U-Net is trained along 763 epochs with ellipsoids and 1742 epochs with real chest CT. Training with the ellipsoids dataset and real chest CT dataset took approximately 20.5 hours and 56 hours, respectively.

\section{Results}

Commonly used quantitative metrics are selected such as peak signal-to-noise ratio (PSNR) and structural similarity (SSIM) to see the quantitative performance of 3D U-NetR. The root mean squared error (RMSE) represents L2 error and is used when calculating PSNR \cite{yuanji2003image}. The definition of the RMSE is given as follows:

\begin{equation}
\label{eq5}
RMSE = \sqrt{\sum_{i}\|\hat{y}_i-y_i\|^2}
\end{equation}

where $\hat{y}$ and $y$ represent the vector forms of the reconstructed and ground truth image, respectively. In addition, sub-index $i$ denotes each pixel. Similarly, PSNR can be defined as:

\begin{equation}
\label{eq6}
PSNR = 20\log{\frac{MAX_i}{RMSE}}
\end{equation}

$MAX_i$ is the maximum value of the image which is 255 for 8-bit images. Even though PSNR is commonly used for image quality assessment, it only calculates pixel-wise cumulative error and does not represent how similar the images are. Therefore, the SSIM is used as a second image quality metric to evaluate the similarity of luminance, contrast, and structure \cite{wang2003multiscale}.

\begin{equation}
\label{eq7}
SSIM(x,y) = \frac{(2\mu_x\mu_y+c_1)(2\sigma_{xy}+c_2)}{(\mu_x^2+\mu_y^2+\sigma_1)(\sigma_x^2+\sigma_y^2+c_2)}
\end{equation}

where $\mu_x$ and $\mu_y$ represent the average of reconstructed and ground truth images, respectively. $\sigma_x^2$ and $\sigma_y^2$ indicate the variance of reconstructed and ground truth images, respectively. In addition, $\sigma_{xy}$ is the covariance of the reconstructed and ground truth image. $c_1$ and $c_2$ constants are calculated based on the dynamic range of images and they are 2.55 and 7.65 correspondingly for 8-bit images.

\subsection{Synthetic Data Reconstruction Results}

\begin{figure*}
  \begin{center}
  \includegraphics[width=4.8in]{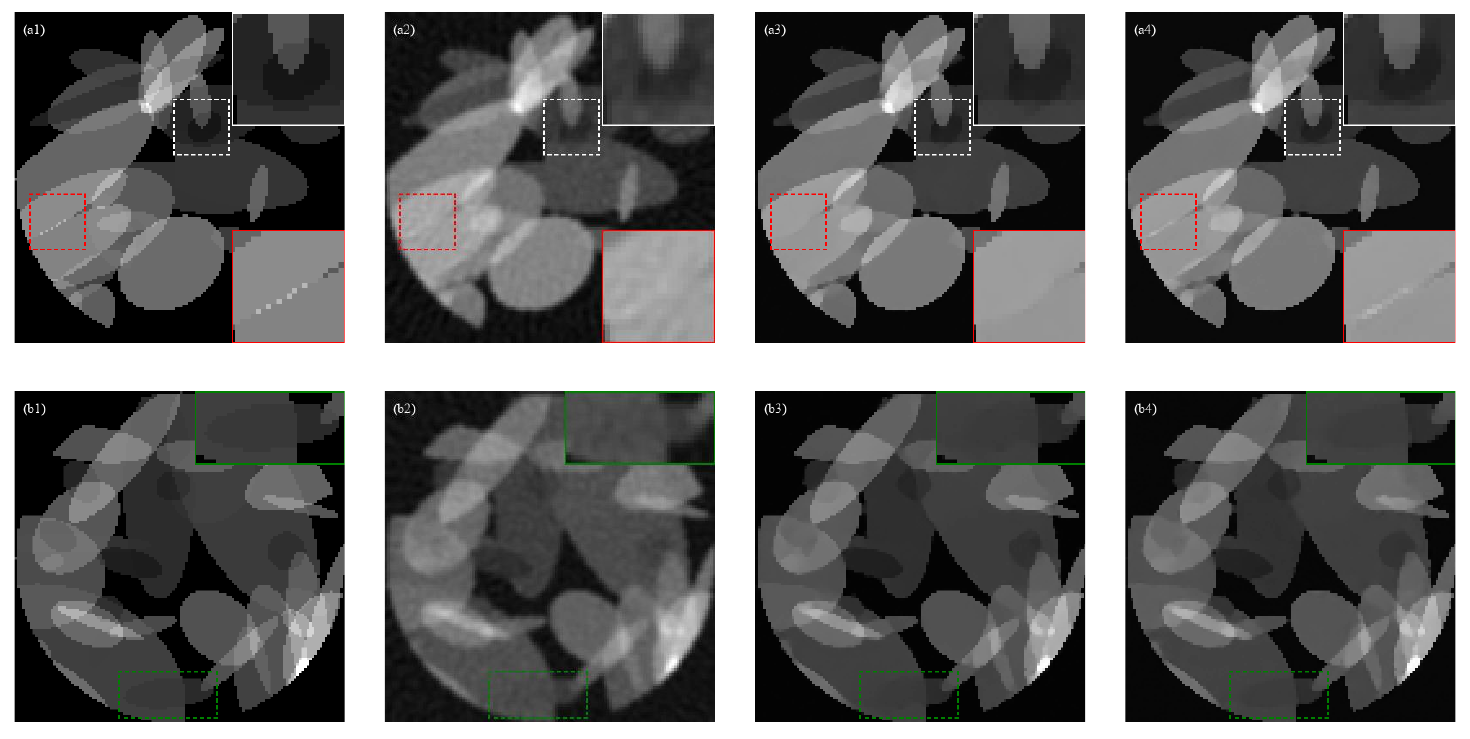}\\
  \caption{2 different synthetic image reconstruction examples. Each row denoted as (a) and (b) represents a different ellipsoid image. Columns (1-4) indicate ground truth, FBP, 2D U-Net, and 3D U-NetR reconstructions, respectively.}\label{fig:synthetic_ex}
  \end{center}
\end{figure*}

The performance of 3D U-NetR is examined with the synthetic dataset in this section. The reconstructed images with the FBP, 2D U-Net, and 3D U-NetR are given in Fig.~\ref{fig:synthetic_ex}. As can be seen from the results, some details lost in FBP images cannot be recovered with 2D U-Net, but with 3D U-NetR. For example, the red zoom window in Fig.~\ref{fig:synthetic_ex} (a1-a4) has a white line-like feature which is missing in 2D U-Net output, but captured by 3D U-NetR. Since 2D U-Net is using only single slice FBP output, some details lost in FBP cannot be recovered in 2D U-Net output. Moreover, it has been observed that the 3D U-NetR reconstructs elliptical edges more smoothly. For instance, when the white zoom windowed parts in Fig.~\ref{fig:synthetic_ex} (a1-a4) are compared, it is observed that the edge of darker ellipse is not preserved in 2D U-Net output while it is recovered more accurately by 3D U-NetR. PSNR and SSIM metric values of test data are given in Table~\ref{tab:synthetic} to show the quantitative performance of the 3D U-NetR. Overall, 3D U-NetR shows slightly higher mean quantitative performance and has a lower standard deviation than 2D U-Net which shows the stability of the 3D U-NetR.

\begin{table}[]
    \centering
    \begin{tabular}{l|c|c|c}
    
         \textbf{Phantom No} & \textbf{SSIM of FBP} & \textbf{SSIM of 2D U-Net} & \textbf{SSIM of 3D U-NetR}\\
         \hline\hline
         Phantom 1 &  72.73$\pm$3.12 & \textbf{97.85$\pm$0.85} & 97.81$\pm$0.71 \\
         Phantom 2 &  69.83$\pm$3.31 & \textbf{98.06$\pm$0.50} & 98.01$\pm$0.47 \\
         Phantom 3 &  73.87$\pm$2.37 & 97.44$\pm$1.20 & \textbf{97.48$\pm$1.03} \\
         Phantom 4 &  71.71$\pm$2.48 & 97.52$\pm$0.96 & \textbf{97.56$\pm$0.80} \\
         Phantom 5 &  72.22$\pm$3.71 & 97.46$\pm$1.36 & \textbf{97.47$\pm$1.19} \\
         Phantom 6 &  75.06$\pm$3.33 & \textbf{97.52$\pm$0.74} & 97.50$\pm$0.67 \\
         Phantom 7 &  73.17$\pm$2.81 & 97.03$\pm$1.29 & \textbf{97.07$\pm$1.06} \\
         Phantom 8 &  74.84$\pm$2.60 & \textbf{97.52$\pm$0.86} & 97.48$\pm$0.73 \\
         Phantom 9 &  71.36$\pm$3.68 & 97.20$\pm$1.38 & \textbf{97.22$\pm$1.18} \\
         Phantom 10 &  73.50$\pm$2.63 & 97.64$\pm$1.20 & \textbf{97.74$\pm$0.98} \\
         Phantom 11 &  71.81$\pm$2.57 & 96.32$\pm$1.45 & \textbf{96.61$\pm$1.11} \\
         Phantom 12 &  73.24$\pm$3.39 & 97.04$\pm$1.12 & \textbf{97.17$\pm$0.94} \\
         Phantom 13 &  73.77$\pm$2.66 & 97.25$\pm$1.06 & \textbf{97.37$\pm$0.81} \\
         Phantom 14 &  72.95$\pm$2.86 & 97.12$\pm$1.35 & \textbf{97.22$\pm$1.12} \\
         Phantom 15 &  76.30$\pm$3.35 & \textbf{97.80$\pm$0.61} & 97.69$\pm$0.64 \\
         Phantom 16 &  75.87$\pm$3.37 & 97.02$\pm$0.75 & \textbf{97.05$\pm$0.62} \\
         Phantom 17 &  73.56$\pm$3.75 & 97.80$\pm$0.68 & \textbf{97.83$\pm$0.65} \\
         Phantom 18 &  70.59$\pm$3.05 & 96.89$\pm$1.41 & \textbf{97.11$\pm$1.09} \\
         Phantom 19 &  74.07$\pm$3.42 & 96.64$\pm$1.11 & \textbf{96.86$\pm$0.86} \\
         Phantom 20 &  73.30$\pm$2.97 & 97.01$\pm$1.16 & \textbf{97.11$\pm$0.90} \\
         \hline
         Average & 73.19$\pm$3.07 & 97.31$\pm$1.05 & \textbf{97.37$\pm$0.88} \\
         
    \end{tabular}
    \begin{tabular}{l|c|c|c}
         \textbf{Phantom No} & \textbf{PSNR of FBP} & \textbf{PSNR of 2D U-Net} & \textbf{PSNR of 3D U-NetR}\\
         \hline\hline
         Phantom 1 &  25.07$\pm$1.09 & \textbf{34.02$\pm$1.58} & 33.99$\pm$1.32 \\
         Phantom 2 &  25.05$\pm$1.38 & 34.43$\pm$1.52 & \textbf{34.52$\pm$1.42} \\
         Phantom 3 &  26.34$\pm$1.26 & 34.64$\pm$2.08 & \textbf{34.68$\pm$1.81} \\
         Phantom 4 &  25.49$\pm$1.44 & 34.21$\pm$2.17 & \textbf{34.29$\pm$1.82} \\
         Phantom 5 &  25.20$\pm$1.76 & \textbf{34.20$\pm$2.75} & 34.13$\pm$2.71 \\
         Phantom 6 &  26.68$\pm$1.60 & \textbf{34.71$\pm$1.63} & 34.70$\pm$1.57 \\
         Phantom 7 &  25.76$\pm$1.37 & 33.70$\pm$2.48 & \textbf{33.71$\pm$2.20} \\
         Phantom 8 &  26.28$\pm$1.25 & \textbf{34.43$\pm$1.64} & 34.31$\pm$1.34 \\
         Phantom 9 &  24.87$\pm$1.30 & 33.50$\pm$1.65 & \textbf{33.63$\pm$1.72} \\
         Phantom 10 &  27.10$\pm$0.99 & 35.40$\pm$2.29 & \textbf{35.53$\pm$2.03} \\
         Phantom 11 &  25.20$\pm$0.87 & 32.87$\pm$1.81 & \textbf{33.04$\pm$1.51} \\
         Phantom 12 &  25.95$\pm$1.42 & 33.48$\pm$1.62 & \textbf{33.71$\pm$1.54} \\
         Phantom 13 &  25.74$\pm$1.22 & 33.99$\pm$1.91 & \textbf{34.06$\pm$1.62} \\
         Phantom 14 &  27.13$\pm$1.23 & 34.86$\pm$1.83 & \textbf{35.01$\pm$1.63} \\
         Phantom 15 &  26.98$\pm$1.59 & \textbf{35.42$\pm$1.68} & 35.30$\pm$1.69 \\
         Phantom 16 &  26.47$\pm$1.36 & 34.03$\pm$1.56 & \textbf{34.07$\pm$1.39} \\
         Phantom 17 &  26.50$\pm$1.48 & 35.06$\pm$1.56 & \textbf{35.11$\pm$1.54} \\
         Phantom 18 &  25.78$\pm$1.49 & 34.01$\pm$2.58 & \textbf{34.12$\pm$2.22} \\
         Phantom 19 &  27.19$\pm$1.17 & 33.73$\pm$1.34 & \textbf{34.04$\pm$1.20} \\
         Phantom 20 &  25.98$\pm$1.21 & 33.83$\pm$1.74 & \textbf{33.86$\pm$1.40} \\
         \hline
         Average & 26.04$\pm$1.32 & 34.23$\pm$1.87 & \textbf{34.29$\pm$1.68} \\
         
    \end{tabular}
    \caption{Quantitative results of the FBP, 2D U-Net and 3D U-NetR with synthetic images}
    \label{tab:synthetic}
\end{table}

\subsection{Real Chest CT Data Reconstruction Results}

Forward propagation of the real chest CT data is done differently from the ellipsoids due to the image size. Even though medical images are patched for training because of GPU limitations, bigger portions of data can be used as a whole in forward propagation thanks to the higher memory capacity of the CPU and RAM. The first thing to note here is the receptive field of network architecture. 3D U-NetR is a deep and complex network that has a receptive field of 140$\times$140$\times$140 voxels. In terms of slices, every slice in forward propagation is affected by the adjacent 70 slices in both directions. Therefore, doing the reconstruction of each patch separately is highly erroneous. Thus, it is decided to separately process each patient's first and last 192 slices and then paste back the first and last 128 slices for forward propagation. Hereby, an interval of 64 slices starting from the middle is used only as padding for other voxels. Still, it is 6 slices less than the ideal value but restrictions on resources enable this as the way to reconstruct with minimum error.

\begin{figure*}
  \begin{center}
  \includegraphics[width=4.5in]{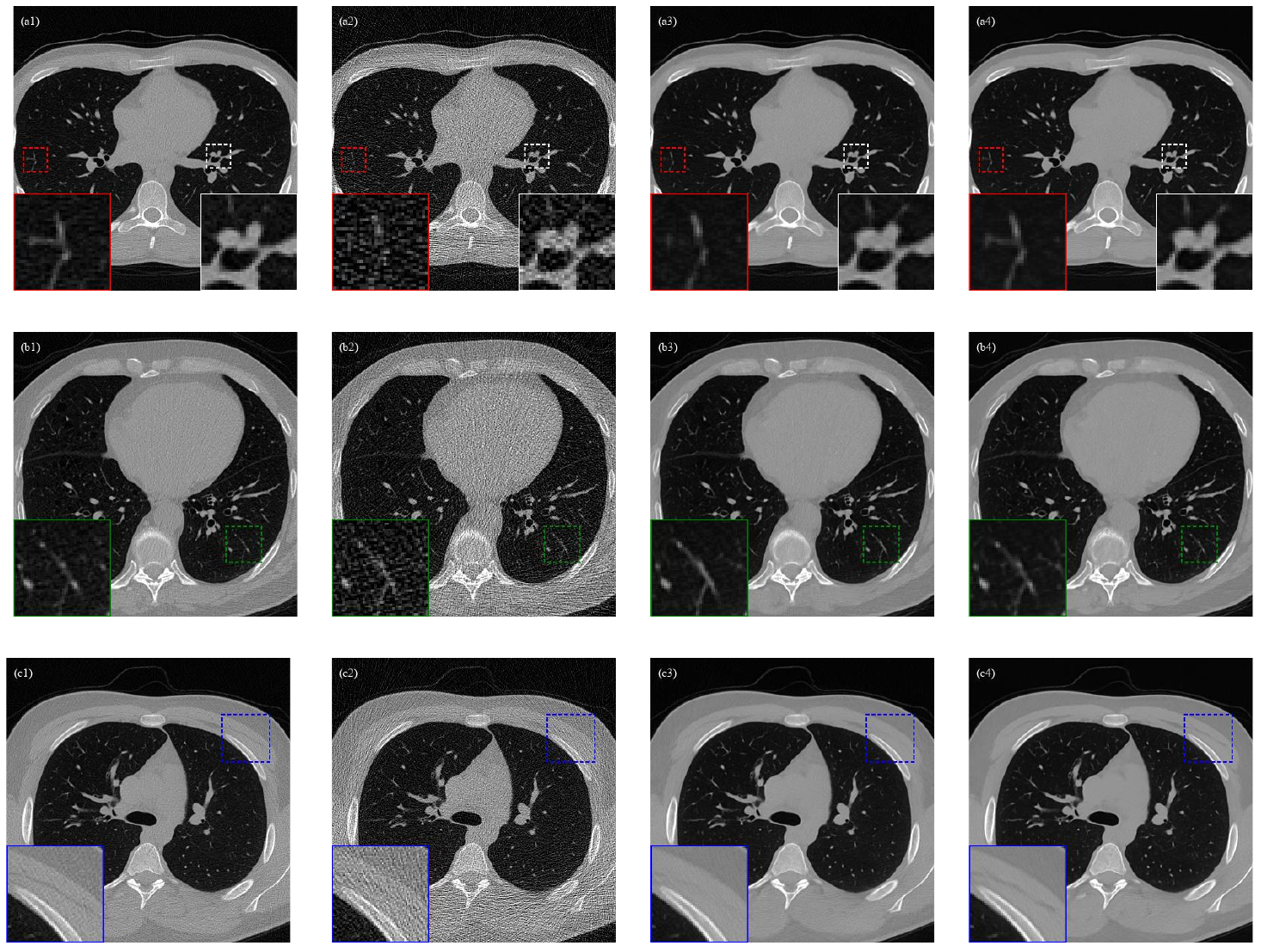}\\
  \caption{3 different real chest CT image reconstruction examples. Each row denoted as (a), (b), and (c) indicates a different real chest CT image. Columns (1-4) represent ground truth, FBP, 2D U-Net, and 3D U-NetR outputs, respectively. The images are displayed using the 0-0.45 normalized window, which corresponds to [-1024, 820] HU.}\label{fig:real_chest_ex}
  \end{center}
\end{figure*}

We further investigated the performance of 3D U-NetR, which has been trained with low and high-dose real chest CT images prepared by the Mayo Clinic. The ill-posed problem with synthetic data is easier than real chest CT images since synthetic data contains images with less resolution. The results for the reconstructed images with the FBP and the forward propagated images with trained 2D U-Net and 3D U-NetR are provided in Fig.~\ref{fig:real_chest_ex}. As can be seen from the figure, 3D U-NetR captures some details that FBP and 2D U-Net cannot reconstruct. Some details lost in the vessels and bone tissue due to noise in low-dose images cannot be obtained with FBP and 2D U-Net. For example, the horizontal vessel in red zoom window in Fig.~\ref{fig:real_chest_ex} (a1-a4) is lost when 2D U-Net is used, but successfully recovered with 3D U-NetR. When the blue zoom window in Fig.~\ref{fig:real_chest_ex} (c1-c4) is examined, it is seen that 3D U-NetR captures a diagonal line-like detail which is missing in 2D U-Net. Overall, 3D U-NetR recovers the details since it takes into account the correlation between slices. The total quantitative results of the 3D U-NetR are displayed in Table~\ref{tab:real_chest} in terms of PSNR and SSIM. The quantitative performance of 3D U-NetR is considerable more than 2D U-Net in every patient for real chest CT data.

\begin{table}[]
    \centering
    \begin{tabular}{l|c|c|c}
         \textbf{Patient No} & \textbf{SSIM of FBP} & \textbf{SSIM of 2D U-Net} & \textbf{SSIM of 3D U-NetR}\\
         \hline\hline
         Patient 1 &  45.33$\pm$8.59 & 80.01$\pm$5.69 & \textbf{81.65$\pm$5.56} \\
         Patient 2 &  35.33$\pm$5.95 & 69.18$\pm$5.93 & \textbf{71.07$\pm$5.84} \\
         Patient 3 &  38.86$\pm$7.97 & 74.55$\pm$6.27 & \textbf{76.20$\pm$6.10} \\
        \hline
        Average & 39.84$\pm$7.50 & 74.58$\pm$5.96 & \textbf{76.31$\pm$5.83} \\
         
    \end{tabular}
    \begin{tabular}{l|c|c|c}
         \textbf{Patient No} & \textbf{PSNR of FBP} & \textbf{PSNR of 2D U-Net} & \textbf{PSNR of 3D U-NetR}\\
         \hline\hline
         Patient 1 & 23.87$\pm$2.04 & 32.97$\pm$1.46 & \textbf{33.82$\pm$1.46} \\
         Patient 2 & 20.75$\pm$1.58 & 30.20$\pm$1.44 & \textbf{30.84$\pm$1.34} \\
         Patient 3 & 22.05$\pm$1.58 & 31.53$\pm$1.12 & \textbf{32.21$\pm$1.11} \\
        \hline
        Average & 22.23$\pm$1.74 & 31.57$\pm$1.34 & \textbf{32.29$\pm$1.30} \\

    \end{tabular}
    \caption{Quantitative performance of the FBP, 2D U-Net and 3D U-NetR with Real CT images}
    \label{tab:real_chest}
\end{table}

\begin{figure*}[htp]
  \begin{center}
  \includegraphics[width=4.45in]{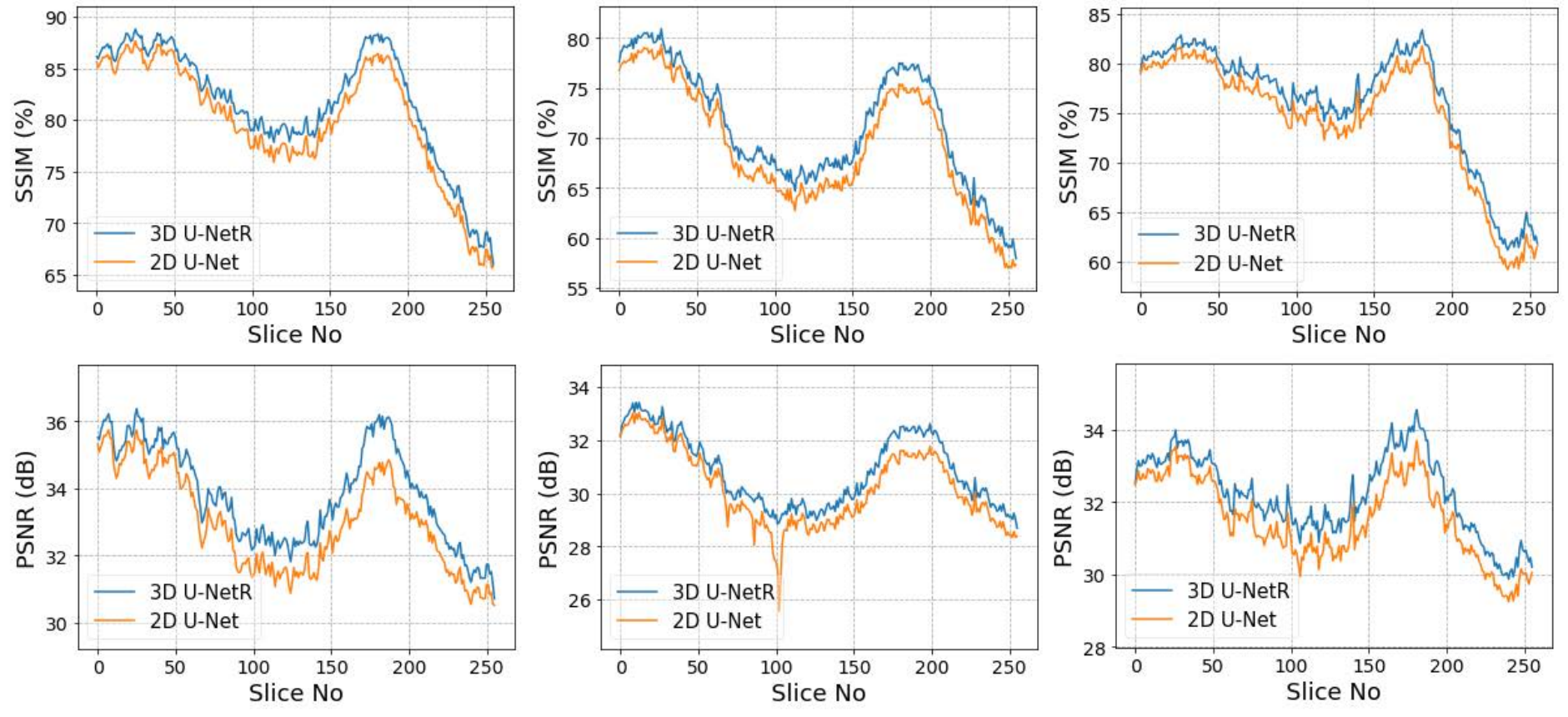}\\
  \caption{PSNR and SSIM values of 2D U-Net (presented with orange line) and 3D U-NetR (presented with blue line) for each slice of 3 test patients whose average scores are given in Table~\ref{tab:real_chest}.}\label{fig:slices}
  \end{center}
\end{figure*}

Although looking at the average PSNR and SSIM values of the reconstructed CT images gives information about performance, it does not does not guarantee the stability of the networks For this reason, it is also important to examine the performances based on slices. Fig.~\ref{fig:slices} shows the SSIM and PSNR values of each slice of the patients one by one. It is observed that 3D U-NetR's superiority in average performance is not due to instant improvements, but generally has better performance in each slice. In addition, 2D U-Net has a high deviation at some points, while 3D U-NetR is more stable since it takes into account the correlation in the third dimension.

\section{Discussion and Conclusion}

In this paper, 3D U-NetR architecture is proposed for CT image reconstruction, inspired by 3D networks previously used for image segmentation. The novel part of 3D U-NetR from other networks that reconstruct CT images in the literature is 3D U-NetR evaluates the image as 3D data and optimizes filter in all three dimensions. The details lost in a 2-dimensional slice can be recovered by examining the third dimension and this has been proven with the prepared experimental setups.

Two different datasets, including synthetic and real chest CT images, are used to validate the model. 3D U-NetR has a difference of 8 dB and 24 percent in terms of PSNR and SSIM, respectively, to the traditional FBP method, and it contains much fewer artifacts visually in the synthetic ellipsoids dataset. When 3D U-NetR is compared with 2D U-Net, the quantitative performance of 3D U-NetR appears to be ahead. In addition, when the images are examined, it is seen that 3D U-NetR reconstructs the edges of the ellipses better and captures some small details which are lost in 2D U-Net's outputs, and this proves that 3D U-NetR is better in synthetic data. The success of 3D U-NetR in sparse view has been shown with the synthetic dataset since it contains a low number of projections.

3D U-NetR has the best quantitative performance among the networks trained with real chest CT images dataset. In addition, it is seen that some details are lost with 2D U-Net and FBP in real chest CT images, as in synthetic data images. However, some vessels in the lung and some tissues in the bone are better recovered with 3D U-NetR. The ability of 3D U-NetR to capture details in vascular and soft tissue is evidence that low-dose CT can become commercially widespread and 3D networks can be used for denoising along with the increasing hardware specs.

The noise model of the synthetic dataset is chosen as Gaussian, which does not mimic perfectly the low-dose CT noise unlike Poisson distribution, for the simplicity in sparse view case. However, Mayo Clinic dataset contains the Poisson noise distribution along with the electronic noise. In addition, the biggest problem in the real chest CT dataset is that the images labeled as ground truth contain a certain amount of noise. The noise in ground truth images causes PSNR and SSIM metrics to be lower than synthetic data. Although PSNR and SSIM metrics are frequently used in the literature, they are not sufficient to emphasize the visual detail differences and quantitative performance in terms of PSNR and SSIM is not fully reliable in medical imaging \cite{pambrun2015limitations}. Therefore, the visually captured details are presented along with quantitative performance in the evaluations.

3D U-NetR gives better results because it takes into account the correlation in the third dimension, but it also has some disadvantages such as high time for training, the limited number of filters, and patched training due to memory limitations.

First of all, since the convolution blocks in the network are 3-dimensional, the number of parameters is approximately three times that of 2-dimensional networks. The high number of parameters causes the network's loss curve to settle with more iterations and require more time for training. Different experimental setups such as residual connection, the filter number, activation function selection, and different configurations of the dataset could not be prepared because the network requires a high time for training. 

Secondly, increasing the number of filters might have an exponential impact on performance compared to 2D convolutions. Although the highest number of possible filters is already preferred in experimental setups, the maximum number of available filters in 3D U-NetR is fewer than 2D U-Net due to memory limitations.

Finally, the high memory consumption of 3D U-NetR obstructs the backpropagation of high-resolution human CT images. Hence, the real chest CT dataset is used by splitting 3D images into patches. Even though the patching operation eliminates the memory limitation, the intersection points of the patches show deformation on direct reconstruction. Thus, patches are designed to have common intersection parts which are equivalent to the half of receptive field of the network as explained in Section 5.2. Overall, the patched training, which is needed because of stricter memory limitations of 3D U-NetR, causes higher calculation time.

The proposed method gives better results with both real and synthetic data compared to its 2-dimensional configuration. Since the scope of the study is the exploration of third-dimensional information with a 3D network, other state of the art networks, such as generative adversarial networks (GAN) and encoder-decoder architecture, are not used in the comparison.

In addition, 3D U-NetR can be applied to different experimental setups containing CT images and different imaging modalities. 3D convolutions can also be used with projection data for improvement in spatial domain. Our future studies will address these topics.

\bibliographystyle{IEEEtran}
\bibliography{references}
\end{document}